\documentclass{article}

\usepackage{microtype}
\usepackage{graphicx}
\usepackage{booktabs}
\usepackage{amsmath,amssymb}
\usepackage{enumitem}
\usepackage{xcolor}
\usepackage{hyperref}

\usepackage[preprint]{icml2026}

\newcommand{\ours}{\textsc{RLScale-Bench}}

\icmltitlerunning{When Does Deep RL Beat Calibrated Baselines?}

\begin{document}

\twocolumn[
\icmltitle{When Does Deep RL Beat Calibrated Baselines? \\
A Benchmark Study on Adaptive Resource Control}

\icmlsetsymbol{equal}{*}

\begin{icmlauthorlist}
\icmlauthor{Guilin Zhang}{wd,gw}
\icmlauthor{Chuanyi Sun}{gw}
\icmlauthor{Kai Zhao}{wd}
\icmlauthor{Xu Chu}{wd}
\icmlauthor{Shahryar Sarkani}{gw}
\icmlauthor{John Fossaceca}{gw}
\end{icmlauthorlist}

\icmlaffiliation{wd}{Workday AI Research}
\icmlaffiliation{gw}{The George Washington University, Washington, DC, USA}

\icmlcorrespondingauthor{Guilin Zhang}{guilin.zhang@gwu.edu}

\icmlkeywords{reinforcement learning, benchmark, adaptive resource control, baseline calibration, reproducibility}

\vskip 0.3in
]

\printAffiliationsAndNotice{}

% ─── Abstract ────────────────────────────────────────────────────────
\begin{abstract}
A properly calibrated rule-based autoscaler can beat every one of six mainstream deep
reinforcement learning (DRL) algorithms on cost across every workload we test---so when,
if ever, does DRL actually help? We study this in \textsc{RLScale-Bench}, a reproducible
benchmark and evaluation protocol for DRL on adaptive resource control, where an agent
allocates compute to a dynamic workload under cost and service-level constraints. The
literature reports conflicting claims about whether model-free RL outperforms well-tuned
rule-based controllers, with single-seed runs, uncalibrated baselines, and inconsistent
training budgets confounding cross-study comparison. We evaluate PPO, DQN, A2C, SAC, TD3,
and DDPG under matched network architectures, training budgets, and reward functions
against a properly calibrated rule-based baseline across six workload patterns and five
seeds (240 runs), instantiate the benchmark on Kubernetes Horizontal Pod Autoscaling, and
probe distribution-shift generalization by training on one workload and deploying on five
shifted distributions.
Three findings challenge common assumptions: (i)~a calibrated rule-based controller achieves
the lowest cost on \emph{all} six workloads and zero constraint violations on steady-state
traffic, though it trails the best RL agents on bursty and flash patterns;
(ii)~discrete-action algorithms outperform continuous-action ones by one to two orders of
magnitude in constraint violations due to action-space mismatch; and (iii)~no single algorithm
dominates across workload types, with rankings shifting by up to four positions between
steady-state and bursty traffic.
On bursty workloads---where RL should shine---PPO reduces constraint violations by 54\%
relative to the calibrated baseline, but only at 24\% higher cost, suggesting that the
bottleneck in RL-based resource control is not algorithm selection but baseline calibration,
reward engineering, and realistic evaluation protocols.

\end{abstract}

% ─── Main body ───────────────────────────────────────────────────────
\section{Introduction}
\label{sec:introduction}

\begin{figure*}[t]
  \centering
  \includegraphics[width=0.92\textwidth]{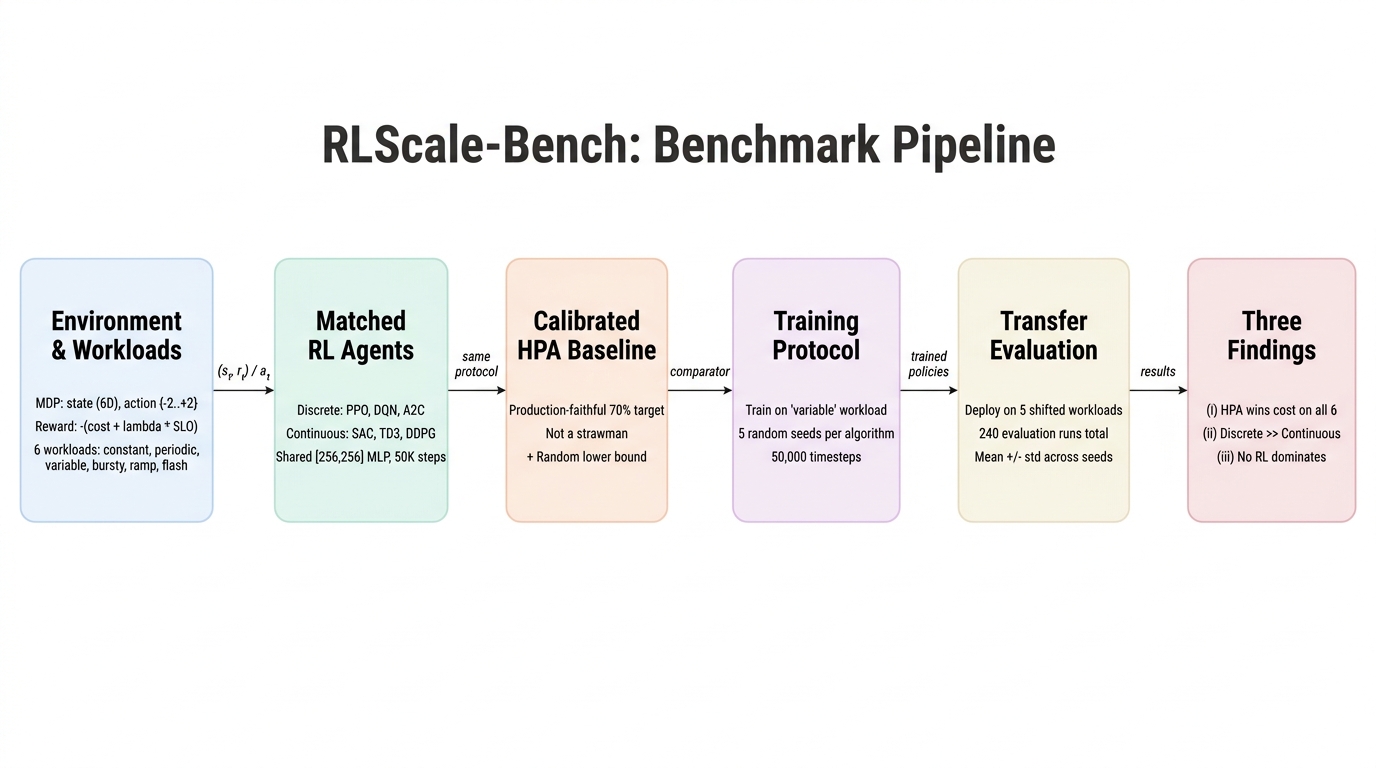}
  \vspace{-0.5em}
  \caption{\ours{} pipeline. The six stages realize our contributions:
  matched RL agents (C1), calibrated HPA baseline (C2), 5-seed training and
  240-run evaluation (C3), deployment on five shifted workloads (C4), and the
  three counter-intuitive findings (C5).}
  \label{fig:pipeline}
  \vspace{-1em}
\end{figure*}

Adaptive resource control---allocating compute resources to a dynamic workload while
respecting cost and service-level constraints---is a canonical decision-making problem
that combines offline training on historical traces with online adaptation to shifting
traffic patterns.
A growing body of work applies deep reinforcement learning (DRL) to this
problem~\citep{rossi2019horizontal,qiu2020firm,wang2022deepscaling,toka2021autoscaling},
typically comparing a single RL algorithm against a simple rule-based controller on one
or two workload patterns.
Yet the conclusions differ sharply across studies: some report that RL improves cost by
30\%, others find that rule-based controllers remain competitive.
This inconsistency hinders progress: without a shared evaluation protocol, practitioners
cannot assess which algorithmic advances are real.

We trace the inconsistency to three gaps in current practice:

\textbf{(1)~Uncalibrated baselines.}
Rule-based controllers such as threshold-driven autoscalers have tunable parameters
(target utilization, cool-down windows) that strongly affect performance.
When RL studies compare against an uncalibrated baseline, apparent improvements may
reflect baseline weakness rather than algorithmic gains~\citep{dulac2021challenges}.

\textbf{(2)~Single-seed reporting.}
DRL training exhibits high variance across random seeds, and many reported
improvements fall within the noise margin of a single method~\citep{henderson2018deep,
islam2017reproducibility,agarwal2021deep}.
Without error bars computed across seeds, benchmark rankings are unreliable.

\textbf{(3)~Narrow workload coverage.}
Most studies evaluate on one or two traffic patterns, typically a diurnal or bursty
trace collected from a single deployment.
Because workload characteristics strongly interact with scaling policies, narrow coverage
yields conclusions that do not transfer to other deployments.

We address these gaps with \textsc{RLScale-Bench}, a benchmark that follows the
reproducible-evaluation principles advocated by~\citet{agarwal2021deep} and extends them
to a real-world decision-making setting.
We instantiate the benchmark on Kubernetes Horizontal Pod Autoscaling---a canonical
adaptive resource control problem deployed in production at large
scale~\citep{k8shpa2024,burns2016borg}---and release an open simulator, trained models,
and evaluation data.
Our contributions are:

\begin{itemize}[leftmargin=*, nosep]
  \item \textbf{C1.} A \textbf{benchmark and evaluation protocol} for adaptive resource control
    that matches network architectures~($[256,256]$ MLP), training budgets~(50K steps),
    and reward functions across PPO, DQN, A2C, SAC, TD3, and DDPG, eliminating confounds
    from implementation choices.
  \item \textbf{C2.} A \textbf{calibrated rule-based baseline} tuned to realistic production
    settings (70\% target utilization), serving as a strong comparator rather than a strawman.
  \item \textbf{C3.} \textbf{Statistically rigorous evaluation} across 5 seeds and 6 workload
    patterns (constant, periodic, variable, bursty, ramp, flash), yielding 240 evaluation
    runs with error bars reported throughout.
  \item \textbf{C4.} A \textbf{distribution-shift generalization study} that trains agents on
    a variable workload and deploys them on five shifted workloads, revealing which algorithms
    adapt and which collapse.
  \item \textbf{C5.} Three \textbf{counter-intuitive findings}: (i)~the calibrated baseline
    achieves lowest cost on all six workloads; (ii)~discrete-action algorithms outperform
    continuous-action ones by orders of magnitude due to action-space mismatch; and
    (iii)~no single RL algorithm dominates across workloads.
\end{itemize}

These findings challenge a common assumption in the decision-making literature that
deep RL straightforwardly outperforms rule-based control.
We argue that progress requires moving beyond algorithm novelty toward reward engineering,
environment calibration, and evaluation protocols that reflect real-world deployment
challenges---themes this benchmark is designed to support.

\section{Related Work}
\label{sec:related}

\paragraph{Decision-making benchmarks for real-world control.}
A growing line of work calls for evaluation protocols that reflect the challenges of
deploying RL in realistic, sequential decision-making settings.
\citet{agarwal2021deep} argue that single-seed results in deep RL benchmarks are
statistically unreliable and propose stratified bootstrap for confidence intervals.
\citet{henderson2018deep} show that hyperparameter choices, random seeds, and
implementation details can flip algorithm rankings, with some reported gains falling
within single-method variance.
\citet{dulac2021challenges} identify weak baselines as a systemic issue across applied RL.
Our benchmark applies these lessons to adaptive resource control, combining matched
training budgets, multiple seeds, and distribution-shift evaluation in a single protocol.

\paragraph{RL for cloud and container resource management.}
RL has been applied to cloud resource management with increasing
sophistication.
\citet{rossi2019horizontal} applied Q-learning to horizontal and vertical container scaling.
\citet{qiu2020firm} proposed FIRM for fine-grained microservice resource management under
SLO constraints.
\citet{wang2022deepscaling} introduced DeepScaling for stable CPU utilization in
large-scale production systems.
\citet{toka2021autoscaling} applied machine learning to Kubernetes edge cluster scaling;
\citet{zhang2025kiss} developed a GPU-aware Kubernetes simulator with PPO-based autoscaling,
demonstrating a 75\% reward improvement over CPU-only baselines;
\citet{zhang2025gpuserverless} addressed adaptive GPU allocation for multi-agent reasoning
in serverless environments;
and \citet{gari2021autoscalingsurvey} survey the broader field of RL-based cloud autoscaling.
These studies typically compare one or two RL algorithms against a lightly-tuned rule-based
controller on a narrow workload range; cross-study comparison is difficult because network
architectures, training budgets, and baseline configurations vary substantially.
Our matched-budget, multi-algorithm, multi-workload protocol is designed to remove these
confounds.

\paragraph{Rule-based autoscaling as a baseline.}
The Kubernetes HPA adjusts replica counts based on observed CPU or custom
metrics~\citep{k8shpa2024}, and KEDA~\citep{keda2024} extends HPA with event-driven
scaling from external sources.
Predictive autoscalers such as the archetype-aware approach of \citet{zhang2025aapa}
augment reactive control with uncertainty-quantified workload forecasts.
While rule-based controllers are often dismissed as ``simple,'' we show that a properly
calibrated HPA is a surprisingly strong comparator---consistent with observations from
\citet{booth2023perils} that reward misdesign and baseline weakness can produce misleading
comparisons in applied RL.

\paragraph{Distribution-shift generalization.}
A central challenge in real-world RL deployment is generalizing from a training
distribution to shifted deployment conditions---a concern that motivates the
offline-to-online RL literature and broader work on distribution shift.
We do not study offline-to-online RL in the strict sense (initializing from a fixed
offline dataset and fine-tuning online); instead, we evaluate distribution-shift
generalization by training each agent online in simulation on a single workload
(\emph{variable}) and deploying it without retraining on five shifted distributions.
Our finding that the best training-time algorithm is rarely the best deployment-time
algorithm connects to broader observations about distribution shift in applied RL.

\paragraph{RL infrastructure and environments.}
Gymnasium~\citep{brockman2016openai} and
Stable-Baselines3~\citep{raffin2021sb3} provide standardized environments and algorithm
implementations.
Microservice benchmarks such as DeathStarBench~\citep{gan2019microservices} provide
realistic workloads but lack integrated RL evaluation frameworks.
\textsc{RLScale-Bench} is, to our knowledge, the first benchmark that combines a realistic
adaptive resource control environment, matched-budget multi-algorithm evaluation, and
explicit distribution-shift evaluation in a single protocol.

\section{Benchmark Design}
\label{sec:method}

Figure~\ref{fig:pipeline} summarizes the benchmark end-to-end; this section
details each stage.

\subsection{Environment}
\label{sec:env}

We instantiate adaptive resource control as a Markov Decision Process (MDP) following
the Gymnasium interface~\citep{brockman2016openai}, using Kubernetes Horizontal Pod
Autoscaling as a concrete testbed.
The agent allocates compute resources (pod replicas) to a service handling a dynamic
request stream, subject to infrastructure cost and service-level constraints.
While we instantiate the benchmark on Kubernetes to ensure production realism---following
prior work on simulation-based evaluation of RL
autoscalers~\citep{zhang2025kiss,zhang2025amp4ec}---the
abstraction applies broadly to cloud scheduling, database provisioning, and
edge inference deployment.

\paragraph{State space.}
The agent observes a 6-dimensional state vector at each decision step:
$s_t = [\text{CPU}_t,\; \text{Mem}_t,\; \text{QPS}_t,\; p95_t,\; \text{ErrRate}_t,\; \text{Replicas}_t]$,
where $\text{CPU}_t \in [0, 100]$ is CPU utilization (\%),
$\text{Mem}_t \in [0, 512]$ is memory usage (MB),
$\text{QPS}_t$ is the current request rate,
$p95_t$ is 95th-percentile latency (ms),
$\text{ErrRate}_t \in [0, 1]$ is the error rate,
and $\text{Replicas}_t \in [1, 10]$ is the current replica count.

\paragraph{Action space.}
The action space is \texttt{Discrete(5)}, representing replica count changes:
$a_t \in \{-2, -1, 0, +1, +2\}$.
This reflects a fundamental constraint of physical resource allocation---replicas are
indivisible units, a property shared by many real-world control problems
(allocation of virtual machines, database shards, or hardware accelerators).
For continuous-action algorithms (SAC, TD3, DDPG), we apply a \texttt{DiscreteToBoxWrapper}
that maps $\text{Box}(-1, 1) \to \text{Discrete}(5)$ via uniform bin edges at $[-0.6, -0.2, 0.2, 0.6]$.

\paragraph{Reward function.}
The reward balances infrastructure cost against SLO compliance:
\begin{equation}
  r_t = -\bigl(\underbrace{c_{\text{rep}} \cdot \text{Replicas}_t}_{\text{cost}}
         + \lambda \cdot \underbrace{\mathbb{1}[\text{SLO violated}]}_{\text{penalty}}\bigr)
  \label{eq:reward}
\end{equation}
where $c_{\text{rep}} = 0.01$ USD per replica per step and $\lambda = 1.0$ controls
the SLO violation penalty.
We normalize rewards to $[-1, 0]$ via min-max scaling based on empirical bounds,
which stabilizes training across all six algorithms.

\paragraph{Workload generator.}
We implement six workload patterns that span the diversity of production traffic
(Table~\ref{tab:workloads}):

\begin{table}[h]
\centering
\small
\caption{Workload types and their characteristics.}
\label{tab:workloads}
\begin{tabular}{@{}llc@{}}
\toprule
Type & Pattern & Range (req/min) \\
\midrule
Constant   & Flat rate + noise        & $100 \pm 10$ \\
Periodic   & Sinusoidal (10-min cycle) & $50$--$200$ \\
Variable   & Random walk              & $30$--$250$ \\
Bursty     & Baseline + Poisson spikes & $50$--$300$ \\
Ramp       & Linear increase          & $50 \to 250$ \\
Flash      & Sudden 3$\times$ spike    & $80 \to 240 \to 80$ \\
\bottomrule
\end{tabular}
\end{table}

\subsection{Algorithms}
\label{sec:algorithms}

We evaluate six DRL algorithms from Stable-Baselines3~\citep{raffin2021sb3}
spanning three families:

\begin{itemize}[leftmargin=*, nosep]
  \item \textbf{On-policy, discrete:} PPO~\citep{schulman2017ppo}, A2C~\citep{mnih2016a2c}
  \item \textbf{Off-policy, discrete:} DQN~\citep{mnih2015dqn}
  \item \textbf{Off-policy, continuous:} SAC~\citep{haarnoja2018sac}, TD3~\citep{fujimoto2018td3}, DDPG~\citep{lillicrap2016ddpg}
\end{itemize}

\noindent All algorithms use:
(i)~identical MLP architecture with two hidden layers of 256 units each,
(ii)~a training budget of 50{,}000 timesteps on the \emph{variable} workload,
and (iii)~5 random seeds per algorithm for statistical robustness.
Algorithm-specific hyperparameters (learning rate, batch size, buffer size) follow
Stable-Baselines3 defaults except where noted in the appendix.

\subsection{Baselines}
\label{sec:baselines}

\paragraph{Calibrated rule-based controller (HPA).}
We implement the production-faithful Kubernetes Horizontal Pod Autoscaler with a 70\%
CPU utilization target: $\text{desired} = \lceil \text{current} \times (\text{CPU} / 70) \rceil$,
clamped to $[1, 10]$ replicas.
The 70\% target follows the default production configuration and provides a built-in
safety margin for bursty traffic.
We emphasize that this baseline is \emph{calibrated}, not a strawman: the target utilization,
clamp bounds, and decision interval match realistic production deployments, enabling a
fair comparison against RL agents.

\paragraph{Random.}
A uniform random policy that selects actions from $\{-2, -1, 0, +1, +2\}$
with equal probability, establishing a lower bound on performance.

\subsection{Evaluation Protocol}
\label{sec:protocol}

Each trained model is evaluated on all six workload types over 240 decision steps
(60 simulated minutes at 15-second intervals).
We report:
\textbf{(1)}~total infrastructure cost (USD),
\textbf{(2)}~total SLO violations (count of steps where latency exceeds threshold
or error rate exceeds 5\%),
and \textbf{(3)}~mean replica count.
All metrics include mean~$\pm$~std across 5~seeds.
For baselines (HPA, Random), we run 5 seeds with different random noise realizations.

\section{Experiments}
\label{sec:experiments}

\subsection{Main Results}
\label{sec:main-results}

Table~\ref{tab:main-results} presents infrastructure costs across all algorithms and workloads.
Our first and most striking finding is that \textbf{HPA achieves the lowest cost on all
six workloads}, outperforming every RL algorithm.
This result holds because HPA scales conservatively---maintaining fewer replicas on average
(2.0--3.0 vs.\ 2.7--3.8 for RL agents)---while still achieving reasonable SLO compliance.

\begin{table*}[t]
\centering
\caption{Total infrastructure cost (USD) across 6 algorithms, 2 baselines, and 6 workload types. Each cell shows mean $\pm$ std over 5 random seeds. \textbf{Bold}: best RL algorithm per workload. \underline{Underline}: best overall. Only algorithms with mean SLO violations $< 100$ are eligible for marking.}
\label{tab:main-results}
\small
\begin{tabular}{lcccccc}
\toprule
Algorithm & Constant & Periodic & Variable & Bursty & Ramp & Flash \\
\midrule
RANDOM & 0.0059 $\pm$ 0.0007 & 0.0059 $\pm$ 0.0007 & 0.0059 $\pm$ 0.0007 & 0.0059 $\pm$ 0.0007 & 0.0059 $\pm$ 0.0007 & 0.0059 $\pm$ 0.0007 \\
HPA & \underline{0.0022 $\pm$ 0.0000} & \underline{0.0025 $\pm$ 0.0000} & \underline{0.0032 $\pm$ 0.0000} & \underline{0.0025 $\pm$ 0.0000} & \underline{0.0033 $\pm$ 0.0000} & \underline{0.0021 $\pm$ 0.0000} \\
\midrule
PPO & 0.0031 $\pm$ 0.0005 & 0.0034 $\pm$ 0.0004 & \textbf{0.0036 $\pm$ 0.0003} & 0.0031 $\pm$ 0.0007 & 0.0039 $\pm$ 0.0003 & \textbf{0.0030 $\pm$ 0.0009} \\
DQN & \textbf{0.0031 $\pm$ 0.0013} & \textbf{0.0034 $\pm$ 0.0008} & 0.0036 $\pm$ 0.0009 & \textbf{0.0029 $\pm$ 0.0013} & 0.0039 $\pm$ 0.0008 & 0.0031 $\pm$ 0.0014 \\
A2C & 0.0032 $\pm$ 0.0008 & 0.0036 $\pm$ 0.0005 & 0.0042 $\pm$ 0.0008 & 0.0038 $\pm$ 0.0011 & \textbf{0.0037 $\pm$ 0.0008} & 0.0036 $\pm$ 0.0007 \\
SAC & 0.0036 $\pm$ 0.0005 & 0.0040 $\pm$ 0.0008 & 0.0041 $\pm$ 0.0009 & 0.0037 $\pm$ 0.0005 & 0.0043 $\pm$ 0.0008 & 0.0039 $\pm$ 0.0002 \\
TD3 & 0.0052 $\pm$ 0.0055 & 0.0052 $\pm$ 0.0055 & 0.0052 $\pm$ 0.0055 & 0.0052 $\pm$ 0.0055 & 0.0052 $\pm$ 0.0055 & 0.0052 $\pm$ 0.0055 \\
DDPG & 0.0011 $\pm$ 0.0000 & 0.0011 $\pm$ 0.0000 & 0.0011 $\pm$ 0.0000 & 0.0011 $\pm$ 0.0000 & 0.0011 $\pm$ 0.0000 & 0.0011 $\pm$ 0.0000 \\
\bottomrule
\end{tabular}
\end{table*}

Table~\ref{tab:slo-results} reports SLO violations.
Here the picture is more nuanced: while HPA achieves zero violations on constant,
periodic, and ramp workloads, it incurs 30.0 violations on bursty traffic---significantly
more than PPO (13.7~$\pm$~5.6) and SAC (8.1~$\pm$~7.5).
This suggests that \textbf{RL's advantage emerges specifically on unpredictable workloads}
where proactive scaling can preempt SLO breaches.

\begin{table*}[t]
\centering
\caption{SLO violations across algorithms and workloads. Each cell shows mean $\pm$ std over 5 seeds. \textbf{Bold}: best RL algorithm. Lower is better; 0 indicates full SLO compliance.}
\label{tab:slo-results}
\small
\begin{tabular}{lcccccc}
\toprule
Algorithm & Constant & Periodic & Variable & Bursty & Ramp & Flash \\
\midrule
RANDOM & 0.1 $\pm$ 0.1 & 152.3 $\pm$ 70.0 & 115.7 $\pm$ 47.9 & 33.4 $\pm$ 40.1 & 216.2 $\pm$ 99.5 & 68.8 $\pm$ 51.3 \\
HPA & 0.0 $\pm$ 0.0 & 0.0 $\pm$ 0.0 & 10.8 $\pm$ 0.0 & 30.0 $\pm$ 0.0 & 0.0 $\pm$ 0.0 & 15.0 $\pm$ 0.0 \\
\midrule
PPO & \textbf{0.0 $\pm$ 0.0} & \textbf{0.0 $\pm$ 0.0} & 12.6 $\pm$ 12.8 & 13.7 $\pm$ 5.6 & \textbf{0.0 $\pm$ 0.0} & 19.1 $\pm$ 17.8 \\
DQN & 0.1 $\pm$ 0.2 & 5.3 $\pm$ 10.6 & 31.2 $\pm$ 31.1 & 17.4 $\pm$ 11.1 & 6.1 $\pm$ 8.3 & 32.5 $\pm$ 31.7 \\
A2C & 0.0 $\pm$ 0.0 & 0.1 $\pm$ 0.2 & 12.9 $\pm$ 12.6 & 12.4 $\pm$ 5.7 & 0.2 $\pm$ 0.3 & 23.3 $\pm$ 21.4 \\
SAC & 0.0 $\pm$ 0.0 & 0.0 $\pm$ 0.0 & \textbf{1.9 $\pm$ 2.5} & \textbf{8.1 $\pm$ 7.5} & 0.0 $\pm$ 0.0 & \textbf{18.9 $\pm$ 24.6} \\
TD3 & 0.5 $\pm$ 0.5 & 452.1 $\pm$ 412.7 & 475.2 $\pm$ 433.8 & 75.0 $\pm$ 68.5 & 593.2 $\pm$ 541.5 & 180.0 $\pm$ 164.3 \\
DDPG & 0.9 $\pm$ 0.0 & 753.5 $\pm$ 0.0 & 791.9 $\pm$ 0.0 & 125.0 $\pm$ 0.0 & 988.6 $\pm$ 0.0 & 300.0 $\pm$ 0.0 \\
\bottomrule
\end{tabular}
\end{table*}

\subsection{Discrete vs.\ Continuous Action Spaces}
\label{sec:discrete-vs-continuous}

\begin{figure}[t]
  \centering
  \includegraphics[width=\columnwidth]{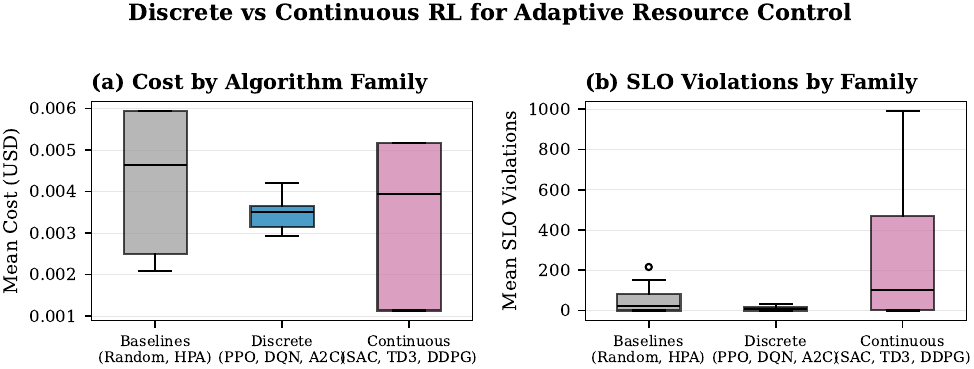}
  \caption{Discrete-action algorithms (PPO, DQN, A2C) achieve dramatically lower SLO
  violations than continuous-action algorithms (SAC, TD3, DDPG).
  The continuous family's median SLO count is $>$100$\times$ higher.}
  \label{fig:family}
\end{figure}

Figure~\ref{fig:family} reveals a dramatic performance gap between algorithm families.
Continuous-action algorithms (SAC, TD3, DDPG) exhibit SLO violations that are
\textbf{one to two orders of magnitude higher} than their discrete-action counterparts
(PPO, DQN, A2C).

TD3 shows extreme variance (std = 4.38 on mean replicas), indicating that some seeds
learn functional policies while others degenerate.
DDPG consistently converges to a single-replica policy across all seeds, resulting in
the lowest cost (\$0.0011) but catastrophic SLO violations (300--989 per workload).

This failure mode arises from the \emph{action-space mismatch}: Kubernetes scaling is
inherently discrete (integer replicas), yet continuous algorithms output real-valued
actions that must be bucketed at bin edges. The discretization aliases the reward signal:
within a bin, small changes to the actor's output produce no change in reward (near-zero
gradient); at bin edges, infinitesimal changes produce discrete jumps (high-variance
gradient). The 50K budget is also short for off-policy continuous algorithms, so part
of the gap may reflect undertraining---a learning-curve ablation is left to future work.

\subsection{Cost-SLO Trade-off}
\label{sec:pareto}

\begin{figure}[t]
  \centering
  \includegraphics[width=\columnwidth]{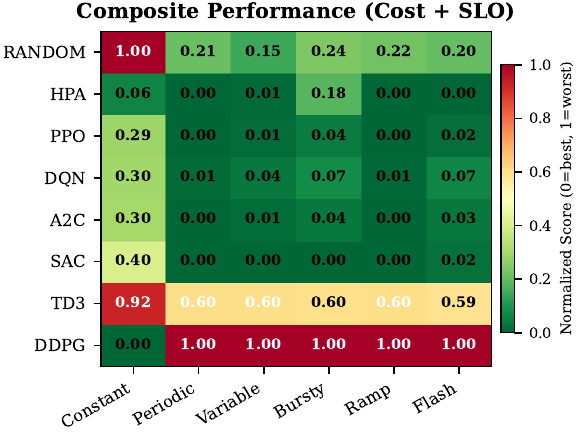}
  \caption{Composite performance heatmap (0 = best, 1 = worst). HPA and SAC
  form the Pareto frontier among viable algorithms. DDPG's degenerate policy
  scores 1.00 on all dynamic workloads.}
  \label{fig:heatmap}
\end{figure}

Figure~\ref{fig:heatmap} shows the composite score (normalized cost + SLO) across
algorithm-workload pairs. HPA achieves the lowest composite scores overall (0.00--0.18);
among RL methods, SAC is strongest on SLO-dominated workloads and PPO on cost-dominated
ones, together forming the cost--SLO frontier analyzed in Appendix~\ref{app:pareto}.
Notably, DDPG achieves a perfect 1.00 (worst) on every non-constant workload due to its
degenerate single-replica policy, confirming that \textbf{cost minimization without SLO
awareness is not a viable strategy}.

\subsection{Ranking Stability Under Workload Shift}
\label{sec:rank-stability}

\begin{figure}[t]
  \centering
  \includegraphics[width=\columnwidth]{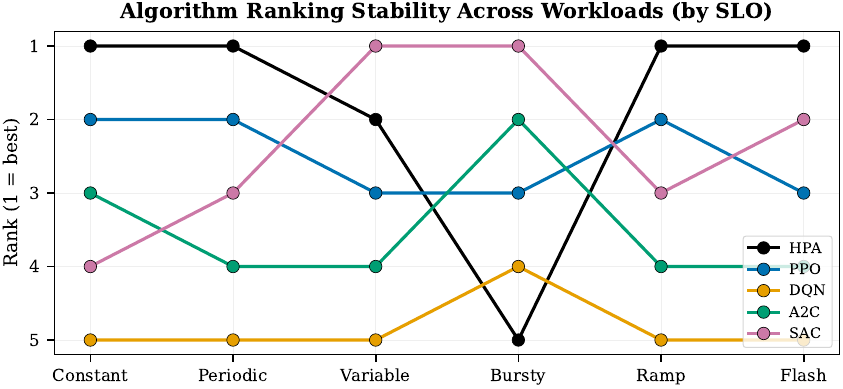}
  \caption{Algorithm ranking (by SLO violations) shifts across workloads.
  No algorithm maintains rank~1 on all patterns. Only viable algorithms shown
  (TD3, DDPG excluded).}
  \label{fig:rank}
\end{figure}

Figure~\ref{fig:rank} tracks how algorithm rankings (by SLO violations) shift across
workload types---a stress test for the common practice of training on one distribution
and deploying on another.
No single algorithm maintains a consistent rank: HPA is rank~1 on constant and periodic
workloads but drops to rank~5 on bursty traffic;
PPO maintains the most stable ranking (rank~2--3);
DQN is consistently the weakest viable algorithm (rank~4--5);
SAC shows the widest rank variance, excelling on variable workloads (rank~1) but
performing poorly on ramp traffic (rank~4).

This instability carries a direct implication for train-to-deploy distribution shift:
\textbf{the best algorithm selected on a training distribution is unlikely to remain
best after workload shift}.
Benchmarks that evaluate only on the training distribution systematically overestimate
deployed performance.

\subsection{Distribution-Shift Generalization}
\label{sec:transfer}

All agents were trained on the \emph{variable} workload---a random-walk trace designed
to expose the policy to diverse conditions---and deployed on the other five workloads
without retraining.
This probes distribution-shift generalization: can a policy trained on one traffic
distribution adapt when deployment conditions shift?

Table~\ref{tab:slo-results} answers this question through the lens of SLO compliance.
PPO, trained on \emph{variable}, achieves 0 violations on \emph{constant}, \emph{periodic},
and \emph{ramp}---it generalizes successfully to these distributions.
On \emph{bursty} and \emph{flash}, however, violations rise to 13.7 and 19.1 respectively,
showing that the policy fails to extrapolate to heavy-tail traffic it did not see during
training.
SAC generalizes more robustly on dynamic workloads (1.9 violations on \emph{variable}
itself, 8.1 on \emph{bursty}) but at the cost of systematically higher resource usage
(3.19--3.78 mean replicas vs.\ 2.68--3.18 for PPO).
DQN and A2C fall between these regimes: both inherit PPO's cost efficiency on steady
workloads but degrade more than PPO on bursty traffic. We call this trade-off the
\textbf{transfer tax}: algorithms that generalize well under distribution shift pay for
robustness with additional resource cost.

Critically, the calibrated rule-based baseline incurs no transfer tax because it does not
train on any distribution---its policy is a fixed function of current CPU utilization.
This explains why a properly tuned rule-based controller beats many RL agents under
distribution shift: it simply does not overfit.

\section{Discussion}
\label{sec:discussion}

\paragraph{Why is the calibrated baseline so competitive?}
A properly tuned rule-based controller is surprisingly hard to beat.
We attribute this to three factors:
(i)~CPU utilization is a strong, low-latency proxy for load in most production workloads,
making reactive scaling effective for steady-state and mildly dynamic traffic;
(ii)~the 70\% target provides a built-in safety margin that prevents constraint violations
under gradual load changes;
(iii)~the rule-based policy does not train on any distribution, so it incurs no
transfer tax when workload shifts.
This echoes observations in broader applied RL that simple, well-tuned baselines are
routinely underestimated~\citep{dulac2021challenges,henderson2018deep}.

\paragraph{When does RL help?}
RL's advantage concentrates on \emph{unpredictable} workloads (bursty, flash) where
proactive scaling---learned from experience---can preempt violations that a reactive
policy cannot avoid.
On bursty traffic, PPO reduces violations by 54\% relative to the calibrated baseline
(13.7 vs.\ 30.0) at 24\% higher cost.
The practical implication: RL is most valuable in deployments where the cost of constraint
violations (revenue loss, user churn) substantially exceeds resource cost.
In cost-dominated settings, the rule-based controller wins.

\paragraph{The action-space mismatch problem.}
TD3 and DDPG fail catastrophically because the discretization wrapper aliases the
reward signal: the actor sees a piecewise-constant reward surface with flat plateaus
inside bins and sharp jumps at bin edges, producing near-zero gradients most of the
time and high-variance estimates at transitions.
SAC partially overcomes this through entropy-regularized exploration but still pays
15--30\% higher cost than discrete alternatives.
The lesson generalizes beyond autoscaling: any decision-making problem with inherently
discrete actions (allocation of indivisible units, discrete network flows, integer
programming) should prefer discrete-action algorithms or develop differentiable abstractions.

\paragraph{Implications for benchmark design in decision-making research.}
Our findings carry three implications for benchmark designers in the
decision-making community:
(1)~\textbf{Calibrate baselines.}
A weak baseline inflates apparent gains; a strong baseline reveals where progress is real.
Benchmarks should document baseline tuning as carefully as RL hyperparameters.
(2)~\textbf{Evaluate under distribution shift.}
Rankings on the training distribution do not predict rankings under deployment shift.
Benchmarks should include held-out distributions as a first-class requirement.
(3)~\textbf{Report error bars across seeds.}
Single-seed runs can flip rankings between discrete and continuous algorithms,
or between PPO and DQN.
Claims based on one seed are unfalsifiable.

\paragraph{Implications for practitioners.}
For production deployments: (i)~always calibrate the rule-based controller before
evaluating RL---many reported gains may reflect baseline weakness rather than algorithmic
superiority; (ii)~prefer discrete-action algorithms (PPO, A2C) for indivisible-unit
resource allocation; (iii)~evaluate on a diversity of workload patterns, because
algorithm rankings are workload-dependent.

\paragraph{Limitations.}
The benchmark uses a simulated environment calibrated to real Kubernetes metrics from a
Kind cluster.
This choice enables reproducibility and scale (240 runs), but may not capture all dynamics
of production clusters (network latency under load, node failures, multi-tenancy).
The training budget (50K steps) is modest; longer training may narrow the gap between
algorithms, though our matched-budget design ensures fair cross-algorithm comparison.
We evaluate single-service scaling; multi-service and cluster-level scaling remain open
challenges that we plan to address in future work via learned graph-based dynamics models.

\section{Conclusion}
\label{sec:conclusion}

We introduced \textsc{RLScale-Bench}, a reproducible benchmark and evaluation protocol
for deep reinforcement learning on adaptive resource control, instantiated on Kubernetes
Horizontal Pod Autoscaling.
A systematic comparison of six algorithms across six workload patterns and five random
seeds reveals that (1)~a calibrated rule-based controller remains the most cost-effective
solution; (2)~discrete-action algorithms outperform continuous-action ones by orders of
magnitude due to action-space mismatch; and (3)~no single algorithm dominates across
workloads, with rankings shifting by up to four positions under distribution shift.

These findings challenge the prevailing narrative that deep RL straightforwardly
outperforms rule-based control in resource management.
We argue instead that the bottleneck lies in reward engineering, baseline calibration,
and evaluation protocols that reflect real-world deployment challenges---problems the
decision-making community can address through shared benchmarks and higher evaluation
standards.
We release the full benchmark suite, trained models, and evaluation data to support
this goal.\footnote{Code available upon publication.}

\paragraph{Future work.}
We plan to extend \textsc{RLScale-Bench} in three directions:
(i)~learned world models for cascade-aware planning, where an agent rolls out trajectories
in a predicted dynamics model to preempt chain failures before they occur;
(ii)~safe planning via constrained rollouts that guarantee zero violations during
deployment;
and (iii)~graph-structured resource control across multi-service topologies, where scaling
decisions propagate through a dependency graph.

% ─── Acknowledgements ────────────────────────────────────────────────
% Uncomment for camera-ready:
% \section*{Acknowledgements}
% This work was supported by ...

% ─── References ──────────────────────────────────────────────────────
\bibliography{references}
\bibliographystyle{icml2026}

% ─── Appendix ────────────────────────────────────────────────────────
\newpage
\appendix
\section{Hyperparameter Details}
\label{app:hyperparameters}

Table~\ref{tab:hyperparams} lists the full hyperparameter configuration for each algorithm.
All algorithms share the same network architecture and training budget;
algorithm-specific parameters follow Stable-Baselines3~\citep{raffin2021sb3} defaults
except where noted.

\begin{table}[h]
\centering
\small
\caption{Hyperparameter configuration. Shared parameters are listed under ``Common'';
algorithm-specific parameters follow SB3 defaults.}
\label{tab:hyperparams}
\begin{tabular}{@{}ll@{}}
\toprule
\textbf{Parameter} & \textbf{Value} \\
\midrule
\multicolumn{2}{@{}l}{\textit{Common}} \\
Network architecture & MLP $[256, 256]$ \\
Training timesteps & 50{,}000 \\
Training workload & Variable \\
Decision interval & 15 seconds \\
Episode length & 240 steps (60 min) \\
Reward normalization & Min-max to $[-1, 0]$ \\
Random seeds & $\{42, 123, 456, 789, 1024\}$ \\
\midrule
\multicolumn{2}{@{}l}{\textit{PPO}} \\
Learning rate & $3 \times 10^{-4}$ \\
Batch size & 64 \\
Number of epochs & 10 \\
Clip range & 0.2 \\
GAE $\lambda$ & 0.95 \\
\midrule
\multicolumn{2}{@{}l}{\textit{DQN}} \\
Learning rate & $1 \times 10^{-4}$ \\
Buffer size & 50{,}000 \\
Batch size & 64 \\
Target update interval & 500 \\
Exploration fraction & 0.3 \\
Final $\epsilon$ & 0.05 \\
\midrule
\multicolumn{2}{@{}l}{\textit{A2C}} \\
Learning rate & $7 \times 10^{-4}$ \\
Number of steps & 5 \\
Value function coefficient & 0.5 \\
Entropy coefficient & 0.01 \\
\midrule
\multicolumn{2}{@{}l}{\textit{SAC}} \\
Learning rate & $3 \times 10^{-4}$ \\
Buffer size & 50{,}000 \\
Batch size & 256 \\
$\tau$ (soft update) & 0.005 \\
Automatic entropy tuning & Yes \\
\midrule
\multicolumn{2}{@{}l}{\textit{TD3}} \\
Learning rate & $1 \times 10^{-3}$ \\
Buffer size & 50{,}000 \\
Batch size & 256 \\
Policy delay & 2 \\
Target noise & 0.2 \\
\midrule
\multicolumn{2}{@{}l}{\textit{DDPG}} \\
Learning rate & $1 \times 10^{-3}$ \\
Buffer size & 50{,}000 \\
Batch size & 256 \\
$\tau$ (soft update) & 0.005 \\
\bottomrule
\end{tabular}
\end{table}

\section{Environment Calibration}
\label{app:calibration}

The simulated environment is calibrated against metrics collected from a real
Kubernetes cluster running Online Boutique on Kind with Prometheus monitoring.
Key calibration points:

\begin{itemize}[leftmargin=*, nosep]
  \item \textbf{CPU response}: $\text{CPU\%} = 5.0 + (\text{QPS} / \text{replicas}) \times 0.7$,
    calibrated so HPA triggers scale-up at $\sim$100 req/min per replica.
  \item \textbf{Latency model}: $p95 = 50 + (\text{QPS} / \text{replicas})^{1.5} \times 0.08$ ms,
    with exponential degradation under overload.
  \item \textbf{SLO threshold}: $p95 < 500$ ms and error rate $< 5\%$.
  \item \textbf{Cost model}: \$0.01 per replica per decision step (15 seconds),
    approximating on-demand cloud pricing at \$0.04/vCPU-hour.
\end{itemize}

\section{Full Results: Viable Algorithm Comparison}
\label{app:full-results}

\begin{figure}[h]
  \centering
  \includegraphics[width=\columnwidth]{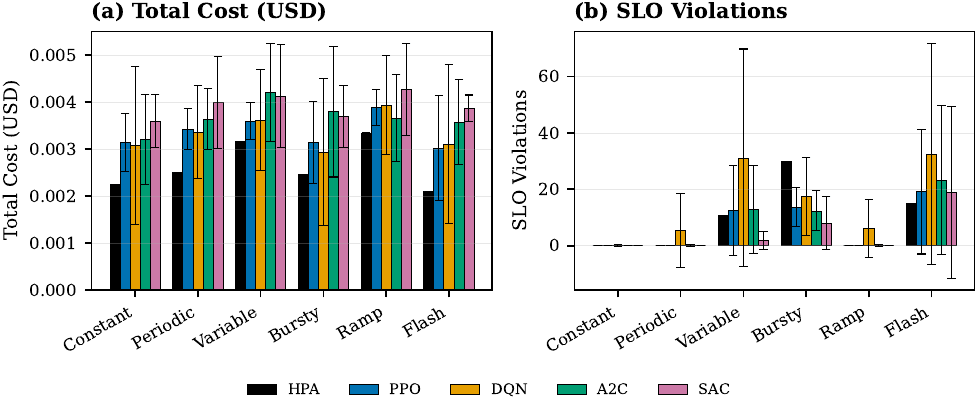}
  \caption{Cost and SLO violations for viable algorithms across all workloads.
  Error bars show 95\% CI over 5 seeds. HPA achieves lowest cost on all workloads.
  RL algorithms show advantage only on bursty/flash SLO compliance.}
  \label{fig:main-viable}
\end{figure}

Figure~\ref{fig:main-viable} shows the complete cost and SLO comparison
for viable algorithms (HPA, PPO, DQN, A2C, SAC) across all six workloads
with 95\% confidence intervals.

\section{Cost-SLO Pareto Analysis}
\label{app:pareto}

\begin{figure}[h]
  \centering
  \includegraphics[width=\columnwidth]{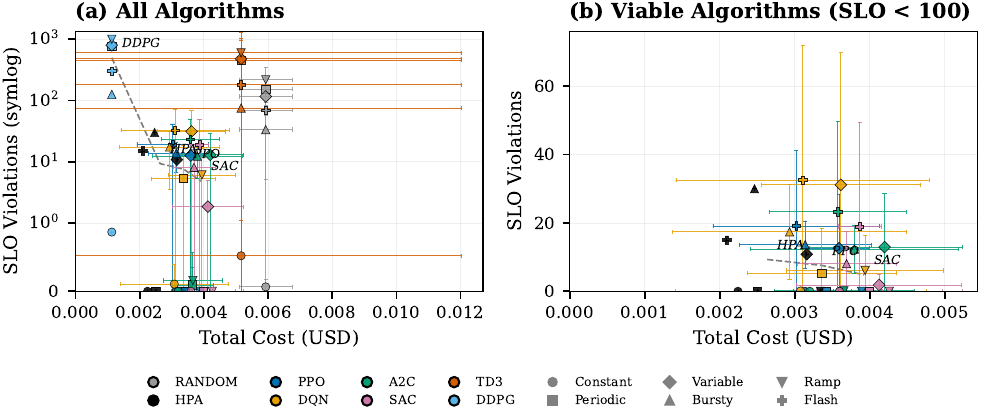}
  \caption{Cost-SLO Pareto front. Left panel: all algorithms (symlog scale).
  Right panel: viable algorithms only. HPA dominates on cost; SAC achieves
  lowest SLO on dynamic workloads.}
  \label{fig:pareto-full}
\end{figure}

Figure~\ref{fig:pareto-full} shows the cost-SLO scatter for all algorithms.
Each point represents one (algorithm, workload) pair.
The Pareto front connects methods that are not dominated on both cost and SLO simultaneously.

\end{document}